\documentclass{article}

\usepackage{microtype}
\usepackage{graphicx}
\usepackage{subfigure}
\usepackage{booktabs} 

\usepackage{amsmath,amsfonts,bm}

\def\eqref#1{equation~\ref{#1}}

\def\1{\bm{1}}

\DeclareMathAlphabet{\mathsfit}{\encodingdefault}{\sfdefault}{m}{sl}
\SetMathAlphabet{\mathsfit}{bold}{\encodingdefault}{\sfdefault}{bx}{n}

\usepackage{hyperref}
\usepackage{url}
\usepackage{amsmath}
\usepackage{pdfpages}
\usepackage{algorithm}
\usepackage{algorithmic}
\usepackage{extpfeil} 
\usepackage{color,xcolor}
\usepackage{epsfig}
\usepackage{graphicx}

\usepackage{adjustbox}
\usepackage{array}
\usepackage{booktabs}
\usepackage{colortbl}
\usepackage{float,wrapfig}
\usepackage{hhline}
\usepackage{multirow}

\usepackage{amsmath,amsfonts,amssymb}
\usepackage{bm}
\usepackage{nicefrac}
\usepackage{microtype}

\usepackage{changepage}
\usepackage{extramarks}
\usepackage{fancyhdr}
\usepackage{lastpage}
\usepackage{setspace}
\usepackage{soul}
\usepackage{xspace}

\usepackage{url}

\usepackage{algorithm, algorithmic}
\usepackage{enumerate}
\usepackage{todonotes} 

\newcolumntype{L}[1]{>{\raggedright\let\newline\\\arraybackslash\hspace{0pt}}m{#1}}
\newcolumntype{C}[1]{>{\centering\let\newline\\\arraybackslash\hspace{0pt}}m{#1}}
\newcolumntype{R}[1]{>{\raggedleft\let\newline\\\arraybackslash\hspace{0pt}}m{#1}}

\newcommand{\ignorethis}[1]{}

\makeatletter
\DeclareRobustCommand\onedot{\futurelet\@let@token\@onedot}
\def\@onedot{\ifx\@let@token.\else.\null\fi\xspace}

\makeatother

\definecolor{MyDarkBlue}{rgb}{0,0.08,1}
\definecolor{MyDarkGreen}{rgb}{0.02,0.6,0.02}
\definecolor{MyDarkRed}{rgb}{0.8,0.02,0.02}
\definecolor{MyDarkOrange}{rgb}{0.40,0.2,0.02}
\definecolor{MyPurple}{RGB}{111,0,255}
\definecolor{MyRed}{rgb}{1.0,0.0,0.0}
\definecolor{MyGold}{rgb}{0.75,0.6,0.12}
\definecolor{MyDarkgray}{rgb}{0.66, 0.66, 0.66}

\usepackage[accepted]{icml2019}

\newcommand{\ourtitle}{CoT: Cooperative Training for Generative Modeling of Discrete Data} 

\icmltitlerunning{\ourtitle}

\begin{document}
 
\twocolumn[
\icmltitle{\ourtitle}



\icmlsetsymbol{equal}{*}

\begin{icmlauthorlist}
\icmlauthor{Sidi Lu}{sjtu}
\icmlauthor{Lantao Yu}{stanford}
\icmlauthor{Siyuan Feng}{sjtu}
\icmlauthor{Yaoming Zhu}{sjtu}
\icmlauthor{Weinan Zhang}{sjtu}
\icmlauthor{Yong Yu}{sjtu}
\end{icmlauthorlist}

\icmlaffiliation{sjtu}{APEX Lab, Shanghai Jiao Tong University, Shanghai, China}
\icmlaffiliation{stanford}{Stanford University, California, USA}

\newtheorem{theorem}{Theorem}

\newtheorem{lemma}{Lemma}

\newtheorem{corollary}{Corollary}

\icmlcorrespondingauthor{Sidi Lu}{steve\_lu@apex.sjtu.edu.cn}
\icmlcorrespondingauthor{Weinan Zhang}{wnzhang@apex.sjtu.edu.cn}

\icmlkeywords{Generative Models, Sequence Modeling, Machine Learning, ICML}

\vskip 0.3in
]



\printAffiliationsAndNotice{} 

\begin{abstract}
  In this paper, we study the generative models of sequential discrete data. 
  To tackle the exposure bias problem inherent in maximum likelihood estimation (MLE), generative adversarial networks (GANs) are introduced to penalize the unrealistic generated samples. 
  To exploit the supervision signal from the discriminator, most previous models leverage REINFORCE to address the non-differentiable problem of sequential discrete data. However, because of the unstable property of the training signal during the dynamic process of adversarial training, the effectiveness of REINFORCE, in this case, is hardly guaranteed. 
  To deal with such a problem, we propose a novel approach called Cooperative Training (CoT) to improve the training of sequence generative models. CoT transforms the min-max game of GANs into a joint maximization framework and manages to explicitly estimate and optimize Jensen-Shannon divergence. 
  Moreover, CoT works without the necessity of pre-training via MLE, which is crucial to the success of previous methods.
  In the experiments, compared to existing state-of-the-art methods, CoT shows superior or at least competitive performance on sample quality, diversity, as well as training stability. 
\end{abstract}

\section{Introduction}

Generative modeling is essential in many scenarios, including continuous data modeling (\emph{e.g.} image generation \citep{goodfellow2014generative,arjovsky2017wasserstein}, stylization \citep{ulyanov2016instance}, semi-supervised classification \citep{radford2015unsupervised}) and sequential discrete data modeling, typically neural text generation \citep{bahdanau2014neural,yu2017seqgan,lu2018neural}. 


For sequential discrete data with tractable density like natural language, generative models are predominantly optimized through Maximum Likelihood Estimation (MLE), inevitably introducing \emph{exposure bias} \citep{ranzato2015sequence}, which results in that given a \textbf{finite} set of observations, the optimal parameters of the model trained via MLE do not correspond to the ones yielding the optimal generative quality. Specifically, the model is trained on the data distribution of inputs and tested on a different distribution of inputs, namely, the learned distribution. This discrepancy implies that in the training stage, the model is never exposed to its own errors and thus in the test stage, the errors made along the way will quickly accumulate.

On the other hand, for general generative modeling tasks, an effective framework, named Generative Adversarial Network (GAN) \citep{goodfellow2014generative}, was proposed to train an implicit density model for continuous data. GAN introduces a discriminator $D_\phi$ parametrized by $\phi$ to distinguish the generated samples from the real ones. As is proved by \citet{goodfellow2014generative}, GAN essentially optimizes an approximately estimated Jensen-Shannon divergence (JSD) between the currently learned distribution and the target distribution. GAN shows promising results in many unsupervised and semi-supervised learning tasks. The success of GAN brings the naissance of a new paradigm of deep generative models, \emph{i.e.} adversarial networks.

However, since the gradient computation requires back-propagation through the generator's output, i.e. the data, GAN can only model the distribution of continuous variables, making it non-applicable for generating discrete sequences like natural language.
Researchers then proposed Sequence Generative Adversarial Network (SeqGAN) \citep{yu2017seqgan}, which uses a model-free policy gradient algorithm to optimize the original GAN objective. With SeqGAN, the expected JSD between current and target discrete data distribution is minimized if the training is perfect. SeqGAN shows observable improvements in many tasks. Since then, many variants of SeqGAN have been proposed to improve its performance. Nonetheless, SeqGAN is not an ideal algorithm for this problem, and current algorithms based on it cannot show stable, reliable and observable improvements that covers all scenarios, according to a previous survey \citep{lu2018neural}. The detailed reasons will be discussed in detail in Section \ref{sec:related-work}.

In this paper, we propose Cooperative Training (CoT), a novel algorithm for training likelihood-based generative models on discrete data by directly optimizing a well-estimated Jensen-Shannon divergence. CoT coordinately trains a generative module $G$, and an auxiliary predictive module $M$, called \emph{mediator}, for guiding $G$ in a cooperative fashion. For theoretical soundness, we derive the proposed algorithm directly from the definition of JSD. We further empirically and theoretically demonstrate the superiority of our algorithm over many strong baselines in terms of generative performance, generalization ability and computational performance in both synthetic and real-world scenarios. 
\section{Background}\label{sec:related-work}
\textbf{Notations.} $P$ denotes the target data distribution. $\theta$ denotes the parameters of the generative module $G$. 
$\phi$ denotes the parameters of the auxiliary predictive mediator module $M$. Any symbol with subscript $_g$ and $_m$ stands for that of the generator and mediator, respectively.
$s$ stands for a complete sample from the training dataset or a generated complete sequence, depending on the specific context. $s_t$ means the $t$-length prefix of the original sequence, \emph{i.e.} an incomplete sequence of length $t$. $x$ denotes a token, and $x_t$ stands for a token that appears in the $t$-th place of a sequence. 
Thus $s_t = [x_0, x_1, x_2, \ldots, x_{t-1}]$ while the initial case $s_0$ is $\emptyset$.

\subsection{Maximum Likelihood Estimation}
Maximum likelihood estimation is equivalent to minimizing the KL divergence using the samples from the real distribution:
\begin{equation}
  \min_\theta \mathbb{E}_{s \sim p_{\text{data}}} \left[ - \log G_\theta(s) \right] ,
\end{equation}
where $G_\theta(s)$ is the estimated probability of $s$ by $G_\theta$ and $p_{\text{data}}$ is the underlying real distribution.

\noindent\textbf{Limitations of MLE.~}
MLE is essentially equivalent to optimizing a directed Kullback–Leibler (KL) divergence between the target distribution $p_{\text{data}}$ and the currently learned distribution $G$, denoted as $KL(p_{\text{data}} \Vert G)$. However, since KL divergence is asymmetric, given \emph{finite} observations this target is actually not ideal. As stated in \citet{arjovsky2017towards}, MLE tries to minimize
\begin{equation}
KL(p_{\text{data}} \Vert G) = \mathop{\sum}_s p_{\text{data}}(s) \mathop{\log} \frac{p_{\text{data}}(s)}{G(s)} .
\end{equation}
\begin{itemize}
	\item When $p_{\text{data}}(s) > 0$ and $G(s) \rightarrow 0$, the KL divergence grows to infinity, which means MLE assigns an extremely high cost to the ``mode dropping'' scenarios, where the generator fails to cover some parts of the data.
	\item When $G(s) > 0$ and $p_{\text{data}}(s) \rightarrow 0$, the KL divergence shrinks to 0, which means MLE assigns an extremely low cost to the scenarios, where the model generates some samples that do not locate on the data distribution.
\end{itemize}
Likewise, optimizing $KL(G \Vert p_{\text{data}})$ will lead to exactly the reversed problems of the two situations. An ideal solution is to optimize a \textbf{symmetrized} and \textbf{smoothed} version of KL divergence, \emph{i.e.} the Jensen-Shannon divergence (JSD), which is defined as
\begin{equation}
JSD(p_{\text{data}} \Vert G) = \frac{1}{2} \big( KL(p_{\text{data}}\Vert M) + KL(G\Vert M) \big),
\end{equation}
where $M = \frac{1}{2}(p_{\text{data}} + G)$.
However, directly optimizing JSD is conventionally considered as an intractable problem. JSD cannot be directly evaluated and optimized since the equally interpolated distribution $M$ is usually considered to be unconstructible, as we only have access to the learned model $G$ instead of $P$.

\subsection{Sequence Generative Adversarial Network}
SeqGAN incorporates two modules, \emph{i.e.} the generator and discriminator, parametrized by $\theta$ and $\phi$ respectively, as in the settings of GAN. By alternatively training these two modules, SeqGAN optimizes such an adversarial target:
\begin{equation}
  \label{eq:gan}
  \mathop{\min}_{\theta} \mathop{\max}_{\phi} \mathbb{E}_{s \sim p_{\text{data}}}\left[\mathop{\log}(D_\phi(s))\right] + \mathbb{E}_{s \sim G_\theta}\left[\mathop{\log}(1 - D_\phi(s))\right] .
\end{equation}
The objectives of generator $G_\theta$ and discriminator $D_\phi$ in SeqGAN can be formulated as:\\
Generator:~~~~\\
\begin{equation}
  \label{eq:REINFORCE}\min_\theta - \mathbb{E}_{s \sim G_\theta} \Big[ \sum^{n}_{t=1}Q_t(s_t, x_t) \cdot \log G_\theta(x_t|s_t) \Big]
\end{equation}
Discriminator:~~~~\\
\begin{equation}
  \label{eq:DISCIMINATION}
  \max_\phi \mathbb{E}_{s \sim p_{\text{data}}}\left[\mathop{\log}(D_\phi(s))\right] + \mathbb{E}_{s \sim G_\theta}\left[\log(1 - D_\phi(s))\right] ,
\end{equation}
where $s \sim G_\theta = [x_1, ..., x_{n}]$ denotes a complete sequence sampled from the generator and the actually implemented action value $Q_t(s_t, x_t) = \mathbb{E}_{s \sim G_\theta(\cdot|s_{t+1})} \left[D_\phi(s)\right]$ is the expectation of the discriminator's evaluation on the completed sequences sampled from the prefix $s_{t+1} = [s_{t}, x_t]$, which can be approximated via Monte Carlo search.

\noindent\textbf{Limitations of SeqGAN \& its Variants.~}
SeqGAN is an algorithm of high variance, which relies on pre-training via Maximum Likelihood Estimation as a variance reduction procedure. During the adversarial epochs, even if with variance reduction techniques such as Actor-Critic methods \citep{sutton1984temporal}, the fact that SeqGAN is essentially based on model-free reinforcement learning makes it a non-trivial problem for SeqGAN to converge well. One consequent result is the ``mode collapse'' problem, which is similar to the original GAN but more severe here. In this case, the learned distribution ``collapses'' towards the minimization of Reverse KL divergence, \emph{i.e.} $KL(G\Vert p_{\text{data}})$, which leads to the loss of diversity of generated samples. In other words, SeqGAN trains the model for better generative quality at the cost of diversity. 


\section{Methodology}
To be consistent with the goal that the target distribution should be well-estimated in both \textbf{quality} and \textbf{diversity} senses, an ideal algorithm for such models should be able to optimize a symmetric divergence or distance. 

\begin{figure}[h]
  \centering 
  \includegraphics[width=1\columnwidth]{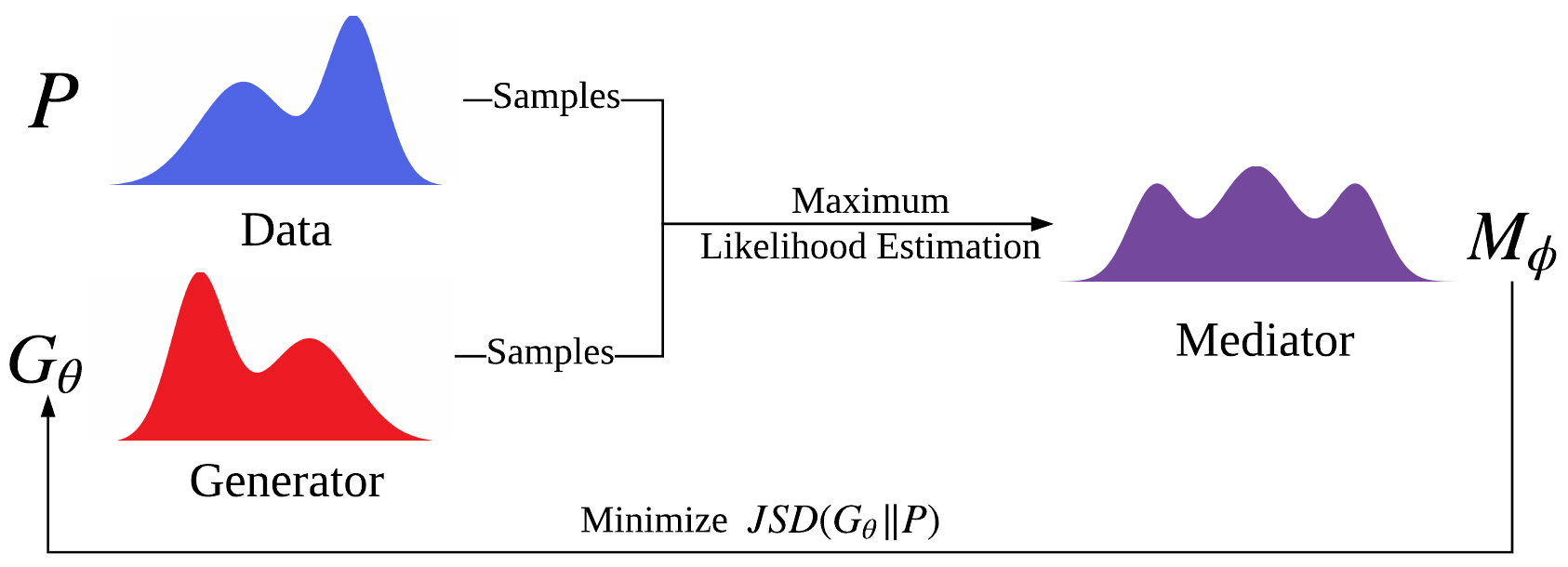}
  \caption{Process of Cooperative Training. }
  \label{fig:cot}
\end{figure}

For sequential discrete data modeling, since the data distribution is decomposed into a sequential product of finite-dimension multinomial distributions (always based on the softmax form), the failures of effectively optimizing JSD when the generated and real data distributions are distant, as discussed in \citet{arjovsky2017wasserstein}, will not appear. 
As such, to optimize JSD is feasible. However, to our knowledge,  no previous algorithms provide a direct, low-variance optimization of JSD. In this paper, we propose Cooperative Training (CoT), as shown in Algorithm~\ref{alg:cot}, to directly optimize a well-estimated JSD for training such models. Figure~\ref{fig:cot} illustrates the whole Cooperative Training process.

\begin{algorithm}[t]
  \caption{Cooperative Training}\label{alg:cot}
  \begin{algorithmic}[1]
    \small
    \REQUIRE
    Generator $G_{\theta}$; mediator $M_\phi$; samples from real data distribution $p_{\text{data}}$; hyper-parameter $N_m$.
    \STATE
    Initialize $G_{\theta}$, $M_\phi$ with random weights $\theta, \phi$.
    \REPEAT
    \FOR{$N_m$ steps}
    \STATE
    Collect two equal-sized mini-batch of samples \{$s_g$\} and \{$s_p$\} from $G_\theta$ and $p_{\text{data}}$, respectively
    \STATE
    Mix \{$s_g$\} and \{$s_p$\} as \{$s$\}
    \STATE
    Update mediator $M_\phi$ with \{$s$\} via Eq.~({\ref{eq:cot_obj_m}})
    \ENDFOR
    \STATE
    Generate a mini-batch of sequences $\{s\} \sim G_\theta$
    \STATE
    Update generator $G_\theta$ with \{$s$\} by applying Eq.~(\ref{eq:simplified_g_grad}) 
    \UNTIL{CoT converges}
  \end{algorithmic}
\end{algorithm}
\subsection{Algorithm Derivation}
\subsubsection{The Objective for Mediator}
Each iteration of Cooperative Training mainly consists of two parts.  
The first part is to train a \emph{mediator} $M_\phi$, which is a density function that estimates a mixture distribution of the learned generative distribution $G_\theta$ and target latent distribution $p_{\text{data}}$ as
\begin{equation}
  M_\phi \simeq \frac{1}{2}(p_{\text{data}} + G_\theta). 
\end{equation}
Since the mediator is only used as a density \textbf{prediction} module during training, the directed KL divergence is now greatly relieved from so-called exposure bias for optimization of $M_\phi$. Denote $\frac{1}{2}(p_{\text{data}} + G_\theta)$ as $M^*$, we have:
\begin{lemma}[Mixture Density Decomposition]
  \label{lemma:mdd}
  \begin{align}
    &\nabla_\phi J_m(\phi) \nonumber \\
    =& \nabla_\phi KL(M^* \Vert M_\phi) \nonumber\\
     =& \nabla_\phi \mathop{\mathbb{E}}_{s \sim M^*} \Big[\mathop{\log}\frac{M^*(s)}{M_\phi(s)}\Big] \nonumber\\
     =& \nabla_\phi \Big( - \mathop{\mathbb{E}}_{s \sim M^*}[\mathop{\log}M_\phi(s)] \Big) \nonumber\\
    =& \nabla_\phi \frac{1}{2}\Big( \mathop{\mathbb{E}}_{s \sim G_\theta}[-\mathop{\log}(M_\phi(s))] + \mathop{\mathbb{E}}_{s \sim p_{\text{data}}}[-\mathop{\log}(M_\phi(s))] \Big)
  \end{align}
\end{lemma}
By Lemma~\ref{lemma:mdd}, for each step, we can simply mix balanced samples from training data and the generator, then train the mediator via Maximum Likelihood Estimation with the mixed samples. The objective $J_m(\phi)$ for the mediator $M$ parameterized by $\phi$ therefore becomes
\begin{align}
  \label{eq:cot_obj_m}
  J_m(\phi) &= \frac{1}{2}\Big( \mathop{\mathbb{E}}_{s \sim G_\theta}[-\mathop{\log}(M_\phi(s))] + \mathop{\mathbb{E}}_{s \sim p_{\text{data}}}[-\mathop{\log}(M_\phi(s))] \Big).
\end{align}

The training techniques and details will be discussed in Section~\ref{sec:exp}.

After each iteration, the mediator is exploited to optimize an estimated Jensen-Shannon divergence for $G_\theta$:
\begin{align}
& J_g(\theta) \nonumber \\
=&   -\hat{JSD}(G_\theta \Vert p_{\text{data}})   \nonumber \\
=&   -\frac{1}{2} \big[KL(G_\theta \Vert M_\phi) + KL(p_{\text{data}} \Vert M_\phi) \big] \nonumber \\
=&   -\frac{1}{2}\mathop{\mathbb{E}}_{s \sim G_\theta} \Big[ \mathop{\log}\frac{G_\theta(s)}{M_\phi(s)} \Big] -\frac{1}{2}\mathop{\mathbb{E}}_{s \sim p_{\text{data}}} \Big[\mathop{\log}\frac{p_{\text{data}}(s)}{M_\phi(s)}\Big] 
\end{align}
When calculating $\nabla_\theta J_g(\theta)$, the second term has no effect on the final results. Thus, we could use this objective instead:
\begin{equation}
    J_g(\theta) =  -\frac{1}{2}\mathop{\mathbb{E}}_{s \sim G_\theta}\Big[\mathop{\log}\frac{G_\theta(s)}{M_\phi(s)}\Big].
\label{eq:original}
\end{equation}

\subsubsection{Generator Objective and Markov Backward Reduction}
For any sequence or prefix of length $t$, we have:
\begin{lemma}[Markov Backward Reduction]
\label{lemma:mbr}
{
\begin{align}
  &  -\frac{1}{2}\mathop{\mathbb{E}}_{s_t \sim G_\theta}\Big[\mathop{\log}\frac{G_\theta(s_t)}{M_\phi(s_t)} \Big] \nonumber\\
  \label{eq:final}
  =& -\frac{1}{2}\mathop{\mathbb{E}}_{s_{t-1} \sim G_\theta}\Big[\mathop{\sum}_{s_t}G_\theta(s_t \vert s_{t-1}) \log\frac{G_\theta(s_t \vert s_{t-1})}{M_\phi(s_t \vert s_{t-1})} \Big] \nonumber \\
  &-\frac{1}{2} \mathop{\mathbb{E}}_{s_{t-1} \sim G_\theta}\Big[\mathop{\log}\frac{G_\theta(s_{t-1})}{M_\phi(s_{t-1})}\Big] .
\end{align}}
\end{lemma}
The detailed derivations can be found in the supplementary material. Note that Lemma~\ref{lemma:mbr} can be applied recursively. That is to say, given any sequence $s_t$ of arbitrary length $t$, optimizing $s_t$'s contribution to the expected JSD can be decomposed into optimizing the first term of Eq.~(\ref{eq:final}) and solving an isomorphic problem for $s_{t-1}$, which is the longest proper prefix of $s_t$. When $t=1$, since in Markov decision process the probability for initial state $s_0$ is always 1.0, it is trivial to prove that the final second term becomes 0.

Therefore, Eq.~(\ref{eq:original}) can be reduced through recursively applying Lemma~\ref{lemma:mbr}. After removing the constant multipliers and denoting the predicted probability distribution over the action space, \emph{i.e.} $G_\theta(\cdot \vert s_{t})$ and $M_\phi(\cdot \vert s_{t})$, as $\mathop{\pi}_g(s_{t})$ and $\mathop{\pi}_m(s_{t})$ respectively, the gradient $\nabla_\theta J_g(\theta)$ for training generator via Cooperative Training can be formulated as
\begin{equation}
  \label{eq:cot_obj_g}
  J_g(\theta) = 
   \mathop{\sum}_{t=0}^{n-1}  \mathop{\mathbb{E}}_{s_t \sim G_\theta} \left[\pi_g(s_t)^\top(\mathop{\log} \pi_m(s_t) - \mathop{\log} \pi_g(s_t)) \right] .
\end{equation}
For tractable density models with finite discrete action space in each step, the practical availability of this objective's gradient is well guaranteed for the following reasons. 
First, with a random initialization of the model, the supports of distributions $G_\theta$ and $P$ are hardly disjoint. 
Second, the first term of Eq.~(\ref{eq:cot_obj_g}) is to minimize the cross entropy between $G$ and $M^*$, which tries to enlarge the overlap of two distributions.
Third, since the second term of Eq.~(\ref{eq:cot_obj_g}) is equivalent to maximizing the entropy of $G$, it encourages the support of $G$ to cover the whole action space, which avoids the case of disjoint supports between $G$ and $P$.

\subsubsection{Factorizing the Cumulative Gradient Through Time for Improved Training}
Up to now, we are still not free from REINFORCE, as the objective Eq.~(\ref{eq:cot_obj_g}) incorporates expectation over the learned distribution $G_{\theta}$. In this part, we propose an effective way to eventually avoid using REINFORCE.
\begin{align*}
&\nabla_\theta J_g(\theta)\\
=&\nabla_\theta\Bigg(\sum_{t=0}^{n-1}\mathop{\mathbb{E}}_{s_t \sim G_\theta} \left[ \pi_g(s_t)^\top (\log \pi_m(s_t) - \log \pi_g(s_t)) \right] \Bigg)
\end{align*}

For time step $t$, the gradient of Eq.~(\ref{eq:cot_obj_g}) can be calculated as 
\begin{align*}
&\nabla_\theta J_{g, t}(\theta)\\
=&\nabla_\theta \left[ \mathop{\mathbb{E}}_{s_t \sim G_\theta} \pi_g(s_t)^\top(\log \pi_m(s_t) - \log \pi_g(s_t))\right]\\
=&\nabla_\theta \left[ \sum_{s_t} G_\theta(s_t) (\pi_g(s_t)^\top(\log \pi_m(s_t) - \log \pi_g(s_t))) \right]\\
=&\sum_{s_t} \nabla_\theta \left[ G_\theta(s_t) (\pi_g(s_t)^\top(\log \pi_m(s_t) - \log \pi_g(s_t))) \right].
\end{align*}
Let $$L(s_t) = \pi_g(s_t)^\top(\log \pi_m(s_t) - \log \pi_g(s_t)),$$
then
\begin{align*}
  &\nabla_\theta J_{g, t}(\theta) \\
  =&\sum_{s_t}(\nabla_\theta G_\theta (s_t) L(s_t) + G_\theta(s_t) \nabla_\theta L(s_t)) \\
  =&\sum_{s_t}G_\theta(s_t)\left(\nabla_\theta \log G_\theta(s_t) L(s_t) + \nabla_\theta L(s_t)\right) \\
  =&\mathop{\mathbb{E}}_{s_t \sim G_\theta}  \nabla_\theta [\text{stop\_gradient}(L(s_t))\log G_\theta(s_t) + L(s_t)].
\end{align*}

The total gradient in each step consists of two terms. The first term $\text{stop\_gradient}(L(s_t))\log G_\theta(s_t)$ behaves like REINFORCE, which makes the main contribution to the variance of the optimization process. The second non-REINFORCE term is comparatively less noisy, though for the first sight it seems not to work alone.

Considering the effects of the two terms, we argue that they have similar optimization directions (towards minimization of $KL(G_\theta \| M_\phi)$ ). To study and control the balance of the two terms, we introduce an extra hyper-parameter $\gamma \in [0, 1]$, to control the balance of the high-variance first term and low-variance second term. The objective in each time step thus becomes
\begin{align*}
  &\nabla_\theta J^{\gamma}_{g, t}(\theta) \\
  =&\mathop{\mathbb{E}}_{s_t \sim G_\theta} \nabla_\theta \left[ \gamma (\text{stop\_gradient}(L(s_t))\log G_\theta(s_t)) + L(s_t) \right].
\end{align*}

In the experiment part, we will show that the algorithm works fine when $\gamma = 0.0$ and the bias of the finally adopted term is acceptable. In practice, we could directly drop the REINFORCE term, the total gradient would thus become
\begin{equation}
  \nabla_\theta J^{0.0}_{g}(\theta) =\mathop{\sum}_{t=0}^{n-1}\mathop{\mathbb{E}}_{s_t \sim G_\theta} \left[ \nabla_\theta  \pi_g(s_t)^\top\Big(\log \frac{\pi_m(s_t)}{\pi_g(s_t)}\Big) \right].
  \label{eq:simplified_g_grad}
\end{equation}

\subsection{Discussions}
\subsubsection{Connection with Adversarial Training}
The overall objective of CoT can be regarded as finding a solution of
\begin{equation}
  \label{eq:COOPERATIVE_TRAINING}
  \mathop{\max}_{\theta} \mathop{\max}_{\phi} \mathop{\mathbb{E}}_{s \sim p_{\text{data}}}\left[\mathop{\log}(M_\phi(s))\right] + \mathop{\mathbb{E}}_{s \sim G_\theta}\left[\mathop{\log}(M_\phi(s))\right].
\end{equation}
Note the strong connections and differences between the optimization objective of CoT (\ref{eq:COOPERATIVE_TRAINING}) and that of GAN (\ref{eq:gan}) lie in the max-max and minimax operations of the joint objective.

\subsubsection{Advantages over Previous Methods}
CoT has several practical advantages over previous methods, including MLE, Scheduled Sampling (SS) \citep{bengio2015scheduled} and adversarial methods like SeqGAN \citep{yu2017seqgan}. 

First, although CoT and GAN both aim to optimize an estimated JSD, CoT is exceedingly more stable than GAN. This is because the two modules, namely generator and mediator, have similar tasks, \emph{i.e.} to approach the same data distribution as \textbf{generative} and \textbf{predictive} models, respectively. 
The superiority of CoT over inconsistent methods like Scheduled Sampling is solid, since CoT has a systematic theoretical explanation of its behavior. Compared with methods that require pre-training in order to reduce variance like SeqGAN \citep{yu2017seqgan}, CoT is computationally cheaper. More specifically, under recommended settings, CoT has the same order of computational complexity as MLE. 

 
Besides, CoT works independently. In practice, it does not require model pre-training via conventional methods like MLE. This is an important property of an unsupervised learning algorithm for sequential discrete data without using supervised approximation for variance reduction or sophisticated smoothing as in Wasserstein GAN with gradient penalty (WGAN-GP) \citep{gulrajani2017improved}.

\subsubsection{The Necessity of the Mediator}
An interesting problem is to ask why we need to train a mediator by mixing the samples from both sources $G$ and $P$, instead of directly training a predictive model $\hat{P}$ on the training set via MLE. There are basically two points to interpret this.

To apply the efficient training objective Eq.~(\ref{eq:cot_obj_g}), one needs to obtain not only the mixture density model $M=\frac{1}{2}(P + G)$ but also its decomposed form in each time step \emph{i.e.} $M_\phi(s) = \prod_{t=1}^{n} M_\phi(s_t\vert s_{t-1})$, without which the term $\pi_m(s_t)$ in Eq.~(\ref{eq:cot_obj_g}) cannot be computed efficiently. This indicates that if we directly estimate $P$ and compute $M = \frac{1}{2} (G + P)$, the obtained $M$ will be actually useless since its decomposed form is not available.

Besides, as a derivative problem of ``exposure bias'', the model $\hat{P}$  would have to generalize to work well on the generated samples \emph{i.e.} $s \sim G_\theta$ to guide the generator towards the target distribution. Given finite observations, the learned distribution $\hat{P}$ is trained to provide correct predictions for samples from the target distribution $P$. There is no guarantee that $\hat{P}$ can stably provide correct predictions for guiding the generator. Ablation study is provided in the supplementary material.
\begin{table*}[t!]
  \caption{Likelihood-based benchmark and time statistics for synthetic Turing test. `-(MLE)' means the best performance is acquired during MLE pre-training.} 
  \label{tb:synthetic_data}
  \begin{center}
  \begin{small}
  \begin{sc}
  \begin{tabular}{lcccr}
    \toprule
    Model     & NLL$_{oracle}$     & NLL$_{test}$(final/best)  & best NLL$_{oracle}$ + NLL$_{test}$ & time/epoch\\
    \midrule
    MLE & 9.08 & 8.97/7.60 & 9.43 + 7.67 & $\textbf{16.14} \pm \textbf{0.97s}$ \\
    SeqGAN\citep{yu2017seqgan} & 8.68 & 10.10/-(MLE) & - (MLE) & $817.64 \pm 5.41s$\\
    RankGAN\citep{lin2017adversarial} & 8.37 & 11.19/-(MLE) & - (MLE) & $1270 \pm 13.01s$ \\
    MaliGAN\citep{che2017maximum} & 8.73 & 10.07/-(MLE) & - (MLE) & $741.31 \pm 1.45s$\\
    Scheduled Sampling & \multirow{2}{*}{8.89} & \multirow{2}{*}{8.71/-(MLE)} & \multirow{2}{*}{- (MLE)} & \multirow{2}{*}{$32.54 \pm 1.14s$}  \\
    \citep{bengio2015scheduled} \\
    Professor Forcing&  \multirow{2}{*}{9.43} &  \multirow{2}{*}{8.31/-(MLE)} &  \multirow{2}{*}{- (MLE)} &  \multirow{2}{*}{$487.13 \pm 0.95s$}  \\
     \citep{lamb2016professor} \\
    \cmidrule{1-5}
    CoT (ours) & \textbf{8.19} & \textbf{8.03/7.54} & \textbf{8.19 + 8.03} & $53.94 \pm 1.01s$\\
    \bottomrule
    \end{tabular}
    \end{sc}
    \end{small}
    \end{center} 
\end{table*}

\begin{figure*}[t!]
  \centering
  \subfigure[JSD of SeqGAN ]{\includegraphics[height=0.45\columnwidth]{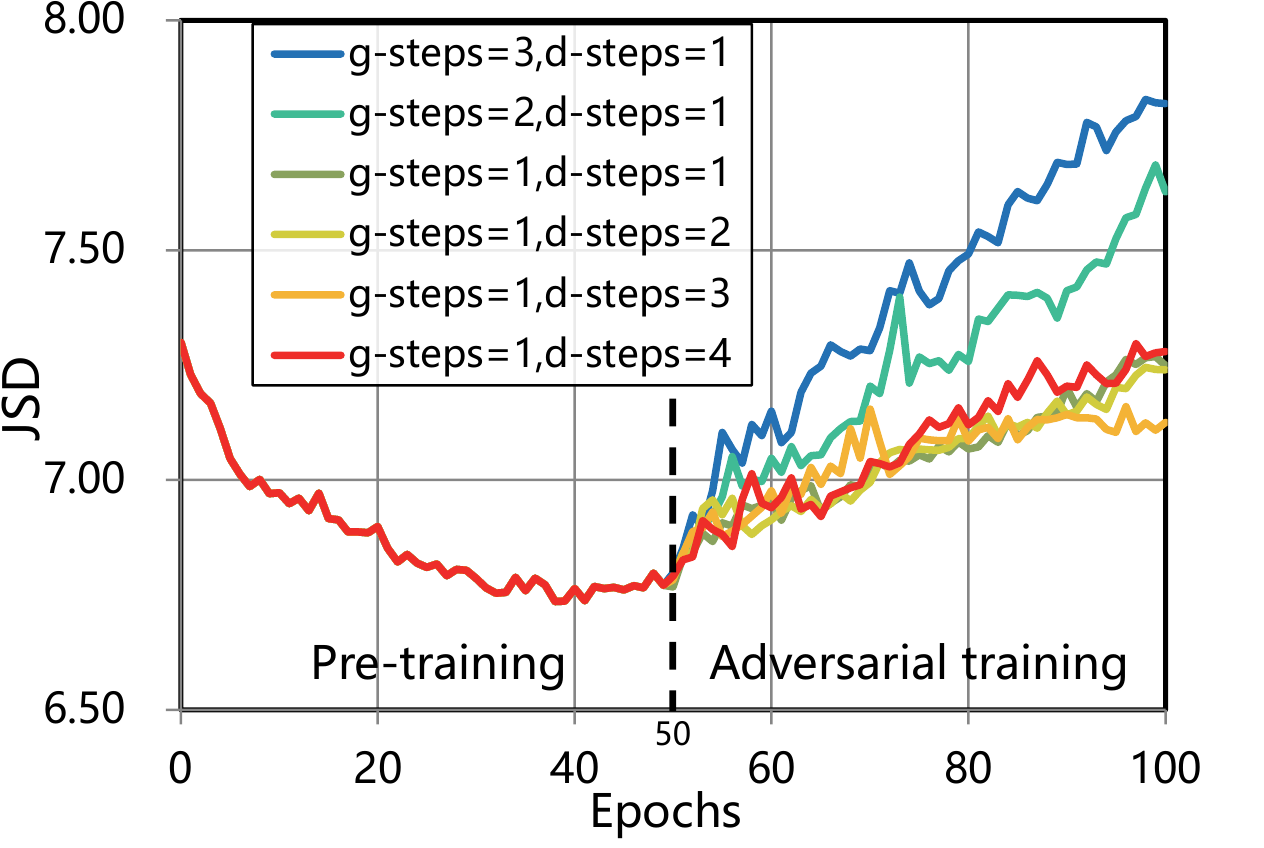}\label{fig:nll_dual}} 
  \subfigure[NLL$_{oracle}$ of CoT]{\includegraphics[height=0.45\columnwidth]{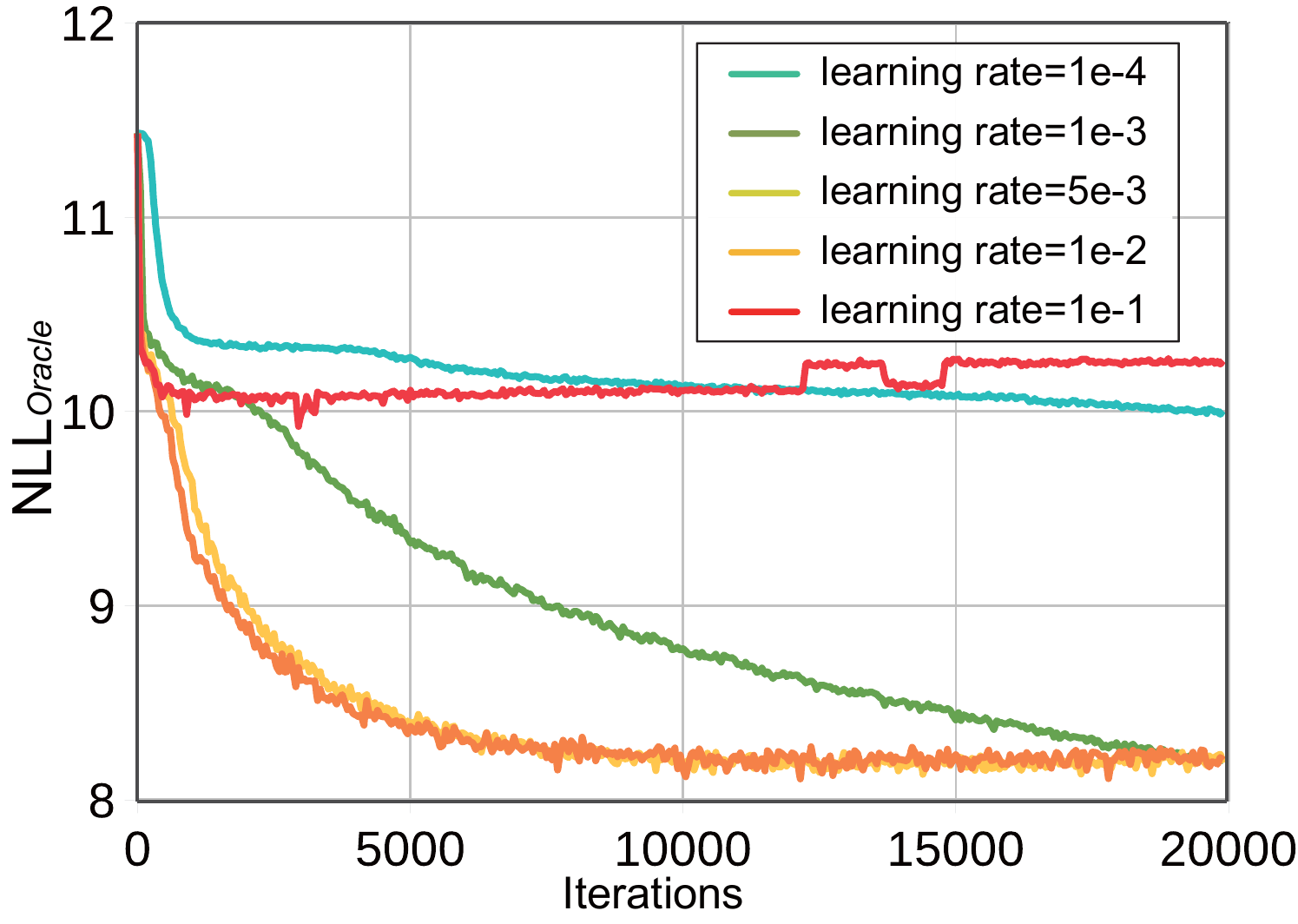}\label{fig:nll_oracle}}
  \subfigure[JSD of CoT]{\includegraphics[height=0.45\columnwidth]{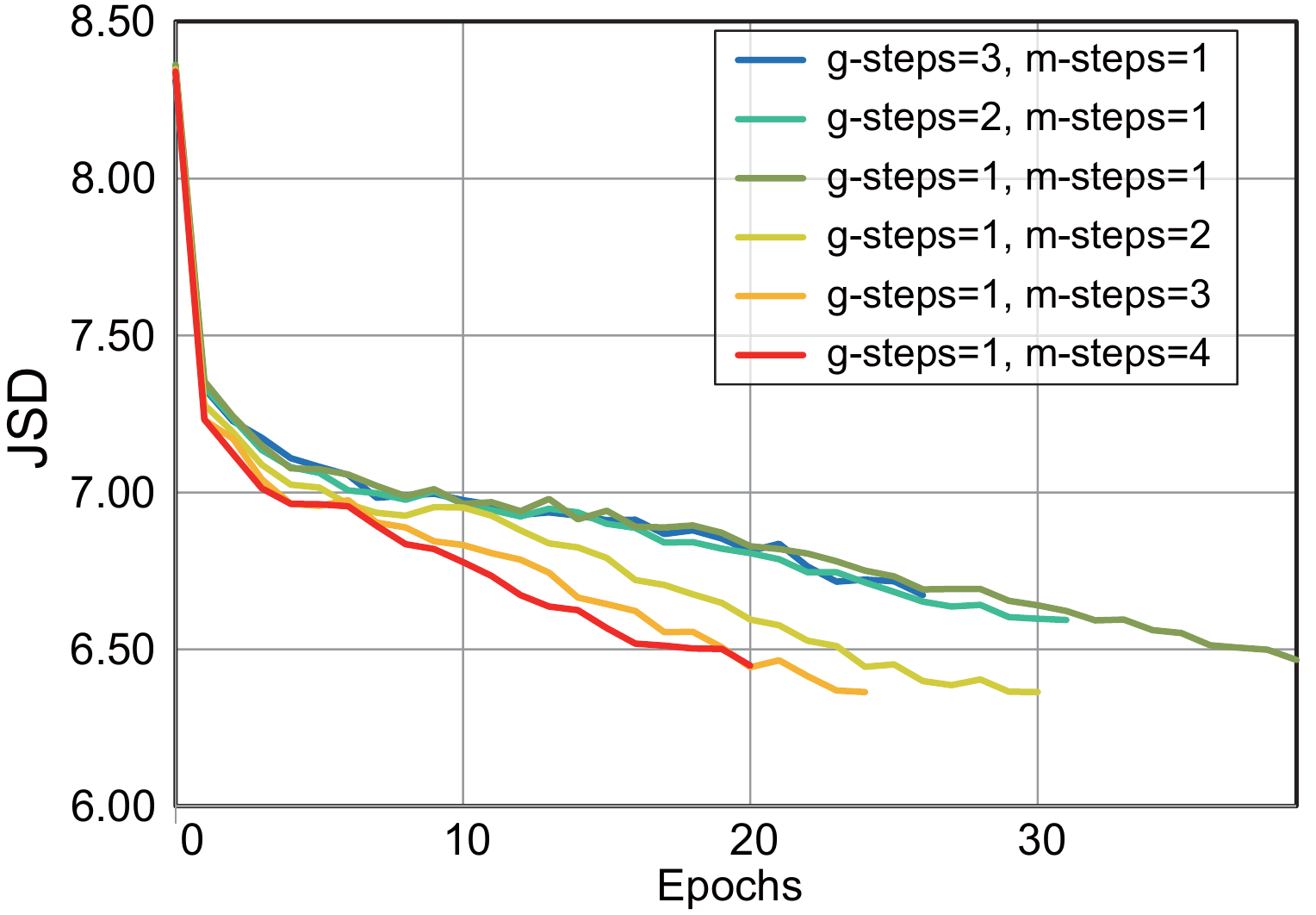}\label{fig:nll_test}}
  \caption{Curves of evaluation on JSD, NLL$_{oracle}$ during iterations of CoT under different training settings. To show the hyperparameter robustness of CoT, we compared it with a typical language GAN \emph{i.e.} SeqGAN \citep{yu2017seqgan}.}
  \label{fig:architecture_robustness}
\end{figure*}
\section{Experiments}\label{sec:exp}
 
\subsection{Universal Sequence Modeling in Synthetic Turing Test}
Following the synthetic data experiment setting in \citet{yu2017seqgan,zhu2018texygen}, we design a synthetic Turing test, in which the negative log-likelihood NLL$_{oracle}$ from an oracle LSTM is calculated for evaluating the quality of samples from the generator. 

Particularly, to support our claim that our method causes little mode collapse, we calculated NLL$_{test}$, which is to sample an extra batch of samples from the oracle, and to calculate the negative log-likelihood measured by the generator. 

We show that under this more reasonable setting, our proposed algorithm reaches the state-of-the-art performance with exactly the same network architecture. Note that models like LeakGAN \citep{guo2017long} contain architecture-level modification, which is orthogonal to our approach, thus will not be included in this part. The results are shown in Table~\ref{tb:synthetic_data}. Code for repeatable experiments of this subsection is provided in supplementary materials.

\subsubsection{Empirical Analysis of Estimated Gradients}
    As a part of the synthetic experiment, we demonstrate the empirical effectiveness of the estimated gradient. During the training of CoT model, we record the statistics of the gradient with respect to model parameters estimated by back-propagating $\nabla_\theta J^{0.0}_{g}(\theta)$ and $\nabla_\theta J^{1.0}_{g}(\theta)$, including the mean and log variance of such gradients. 
    
    We are mainly interested in two properties of the estimated gradients, which can be summarized as:
    \begin{itemize}
        \item \textbf{Bias} Obviously, $\nabla_\theta J^{1.0}_{g}(\theta)$ is exactly the original gradient which is unbiased towards the minimization of Eq.~(\ref{eq:cot_obj_g}). If the estimated gradient $\nabla_\theta J^{0.0}_{g}(\theta)$ is highly biased, the cosine similarity of the average of $\nabla_\theta J^{0.0}_{g}(\theta)$ and $\nabla_\theta J^{1.0}_{g}(\theta)$ would be close to 0.0, otherwise it would be close to 1.0. To investigate this, we calculate the cosine similarity of expected $\nabla_\theta J^{0.0}_{g}(\theta)$ and $\nabla_\theta J^{1.0}_{g}(\theta)$.  \\
        \item \textbf{Variance} We calculate the log variance of $\nabla_\theta J^{0.0}_{g}(\theta)$ and $\nabla_\theta J^{1.0}_{g}(\theta)$ in each dimension, and compute the average log variance of each variance. In the figure, to better illustrate the comparison, we plot the advantage of mean log variance of $\nabla_\theta J^{1.0}_{g}(\theta)$ over $\nabla_\theta J^{0.0}_{g}(\theta)$. If the variance of the estimated gradient is lower, such a statistic would be steadily positive.
    \end{itemize}
    
\begin{figure}[h]
      \centering
      \subfigure[Curves of cosine similarity of averaged $\nabla_\theta J^{0.0}_{g}(\theta) $ and $\nabla_\theta J^{1.0}_{g}(\theta)$ during training.]{\includegraphics[width=0.85\columnwidth]{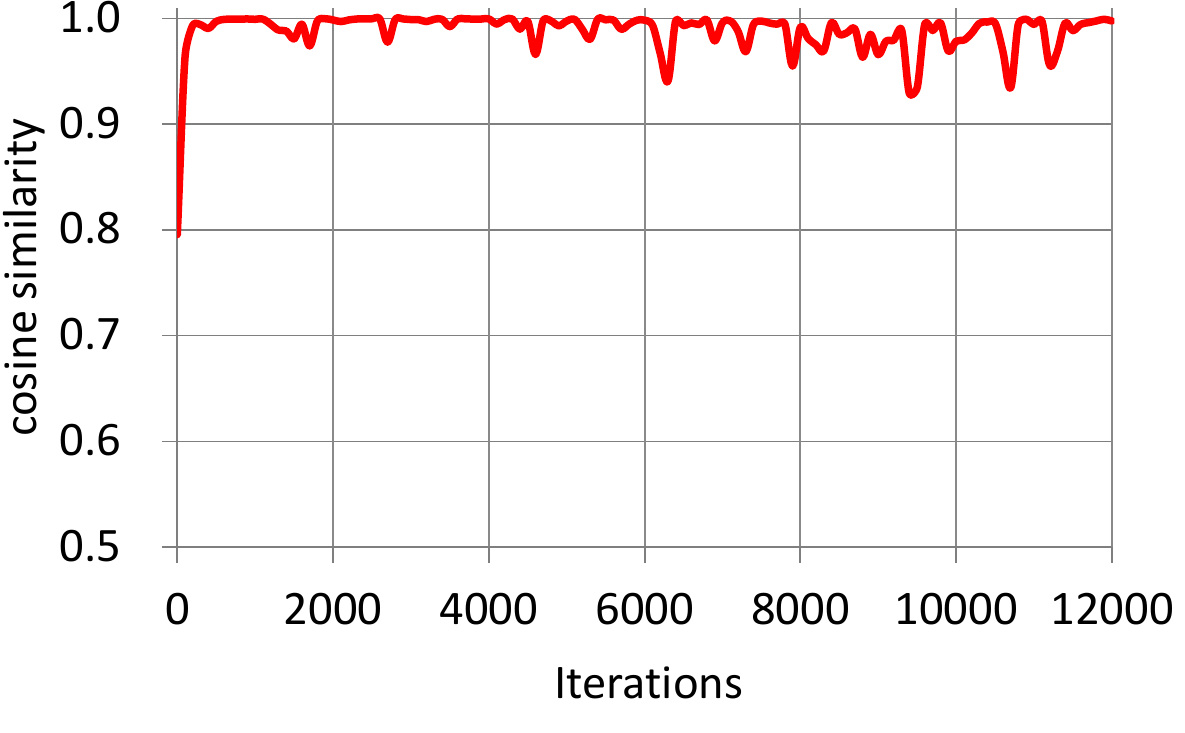}}
      \subfigure[Curves of log variance reduction per dimension of $\nabla_\theta J^{0.0}_{g}(\theta)$ compared to $\nabla_\theta J^{1.0}_{g}(\theta)$]{\includegraphics[width=0.85\columnwidth]{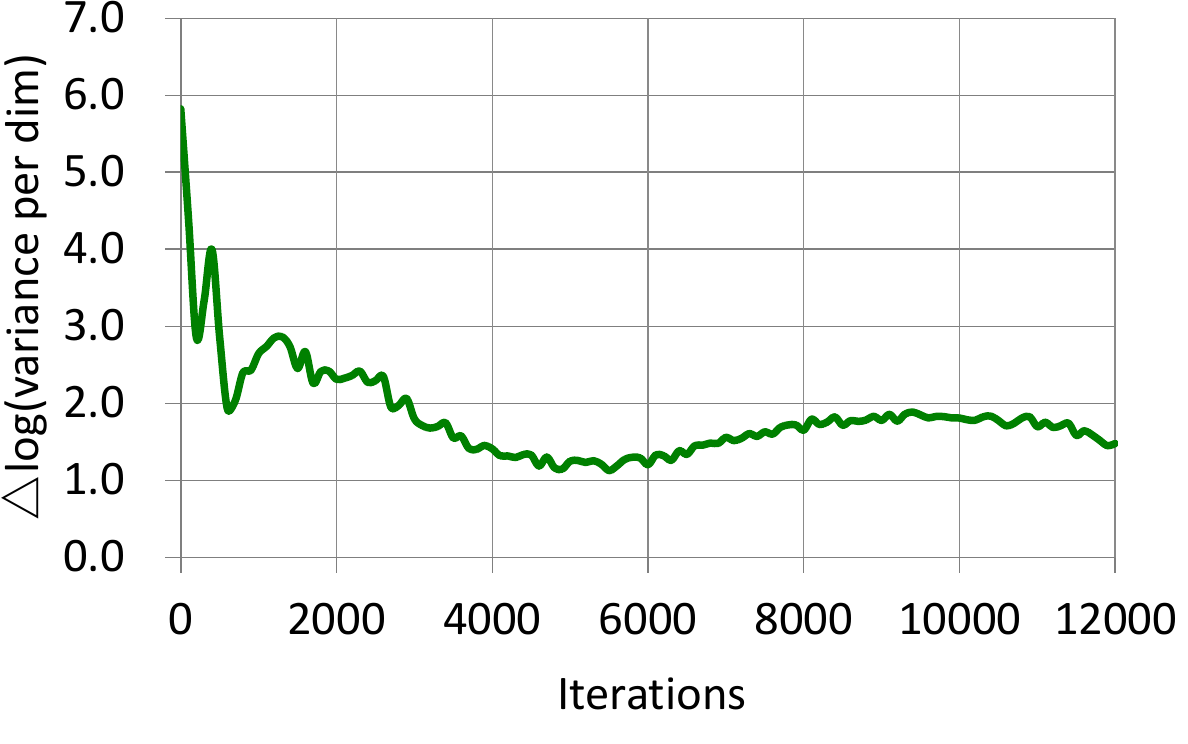}}
      \caption{Empirical study on bias and variance comparison.} 
      \label{fig:grad}  
    \end{figure}

    To calculate these statistics, we sample 3,000 sequences from the generator and calculate the average gradient under each settings every 100 iterations during the training of the model. The results are shown in Figure~\ref{fig:grad}. The estimated gradient of our approach shows both properties of low bias and effectively reduced variance.
    
\subsubsection{Discussion}

  \noindent \textbf{Computational Efficiency}
  Although in terms of time cost per epoch, CoT does not achieve the state-of-the-art, we do observe that CoT is remarkably faster than previous language GANs. Besides, consider the fact that CoT is a sample-based optimization algorithm, which involves time cost in sampling from the generator, this result is acceptable. The result also verifies our claim that CoT has the same order (\emph{i.e.} the time cost only differs in a constant multiplier or extra lower order term) of computational complexity as MLE.

  \noindent \textbf{Hyper-parameter Robustness}
  We perform a hyper-parameter robustness experiment on synthetic data experiment. When compared with the results of similar experiments as in SeqGAN \citep{yu2017seqgan}, our approach shows less sensitivity to hyper-parameter choices, as shown in Figure~\ref{fig:architecture_robustness}. Note that in all our attempts, the curves of the evaluated JSD of SeqGAN fail to converge.
  
\begin{figure}[h]
  \centering
  \subfigure[Curves of $JSD(G\Vert P)$ during training for MLE, SeqGAN and CoT.]{\includegraphics[width=0.85\columnwidth]{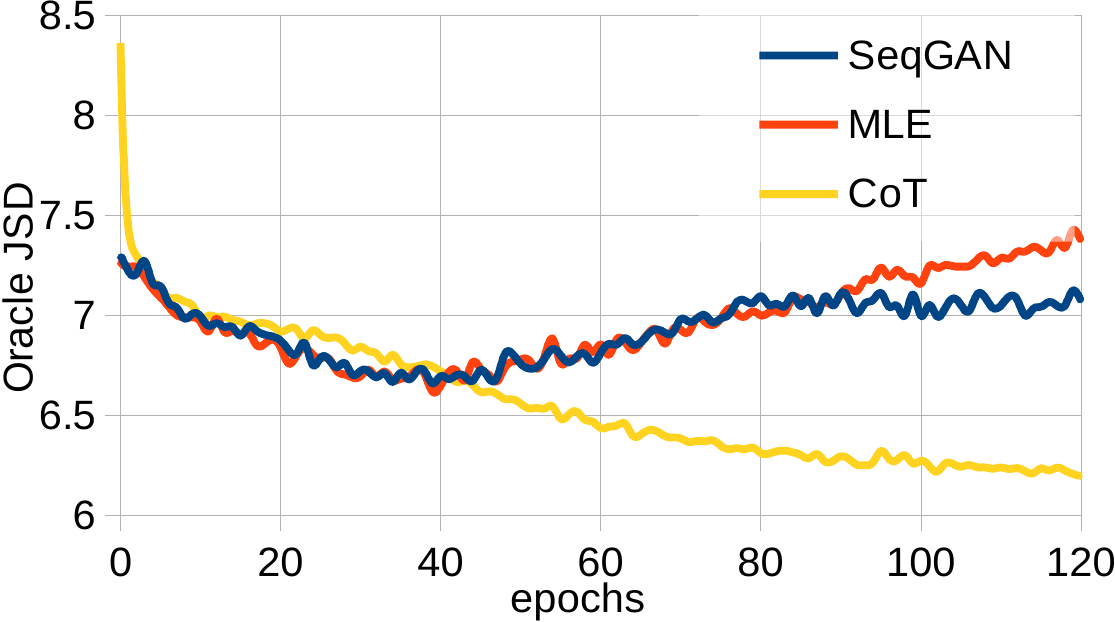}}
  \subfigure[Curves of balanced NLL and real JSD. Both results are from synthetic data experiments.]{\includegraphics[width=0.85\columnwidth]{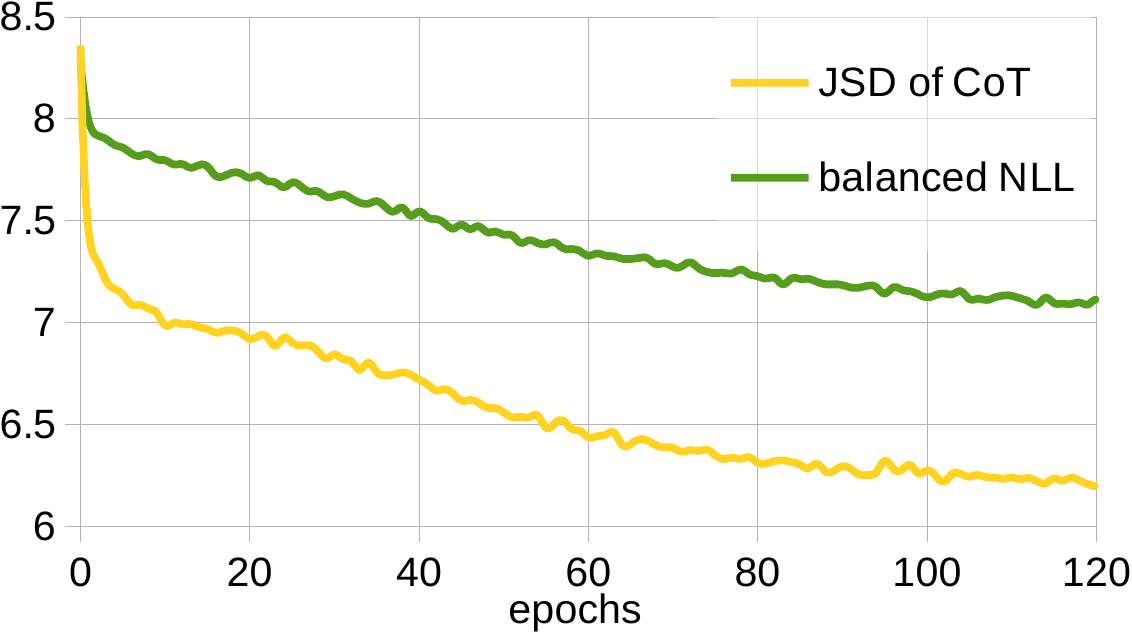}}
  \caption{Training progress curves indicated by different values.} 
  \label{fig:bnll}  
\end{figure}

  \noindent \textbf{Self-estimated Training Progress Indicator}
  Like the critic loss, \emph{i.e.} estimated Earth Mover Distance, in WGANs, we find that the training loss of the mediator (\ref{eq:cot_obj_m}), namely \emph{balanced NLL}, can be a real-time training progress indicator as shown in Figure~\ref{fig:bnll}. Specifically, in a wide range, balanced NLL is a good estimation of real $JSD(G\Vert P)$ with a steady translation, namely, $2~NLL_{balanced} = 2JSD(G\Vert P) + H(G) + H(P)$.
  

\subsection{TextCoT: Zero-prior Long \& Diverse Text Generation}
As an important sequential data modeling task, zero-prior text generation, especially long and diversified text generation, is a good testbed for evaluating the performance of a generative model. 

Following the experiment proposed in LeakGAN \citep{guo2017long}, we choose EMNLP 2017 WMT News Section as our dataset, with maximal sentence length limited to 51. We pay major attention to both \textbf{quality} and \textbf{diversity}. To keep the comparison fair, we present two implementations of CoT, namely CoT-basic and CoT-strong. As for CoT-basic, the generator follows the settings of that in MLE, SeqGAN, RankGAN and MaliGAN. As for CoT-strong, the generator is implemented with the similar architecture in LeakGAN.
 
For quality evaluation, we evaluated BLEU on a small batch of test data separated from the original dataset. For diversity evaluation, we evaluated the estimated Word Mover Distance \citep{kusner2015word}, which is calculated through training a discriminative model between generated samples and real samples with 1-Lipschitz constraint via gradient penalty as in WGAN-GP \citep{gulrajani2017improved}.  To keep it fair, for all evaluated models, the architecture and other training settings of the discriminative models are kept the same.


\begin{table}[t]
  \caption{N-gram-level quality benchmark: BLEU on test data of EMNLP2017 WMT News.\\
  *: Results under the conservative generation settings as is described in LeakGAN's paper.}
  \label{tb:real_quality}
  \begin{center}
  \begin{small}
  \begin{sc}
  \begin{tabular}{lcccr}
    \toprule 
    Model     & BLEU2     & BLEU3 & BLEU4 & BLEU5\\
    \midrule 
    MLE & 0.781 & 0.482 & 0.225 & 0.105 \\
    SeqGAN & 0.731 & 0.426 & 0.181 & 0.096\\
    RankGAN & 0.691 & 0.387 & 0.178 & 0.095 \\
    MaliGAN & 0.755 & 0.456 & 0.179 & 0.088 \\
    LeakGAN* & 0.835 & 0.648 & 0.437 & 0.271 \\
    \cmidrule{1-5}
    CoT-basic & 0.785 & 0.489 & 0.261 & 0.152 \\
    CoT-strong & 0.800 & 0.501 & 0.273 & 0.200\\
    CoT-strong*& \textbf{0.856} & \textbf{0.701} & \textbf{0.510} & \textbf{0.310}\\
    \bottomrule
    \end{tabular}
    \end{sc}
    \end{small}
    \end{center} 
\end{table}
\begin{table}[t]
  \caption{Diversity benchmark: estimated Word Mover Distance (eWMD) and NLL$_{test}$}
  \label{tb:real_diversity}
  \begin{center}
  \begin{small}
  \begin{sc}
  \begin{tabular}{lcccr}
    \toprule
    Model     & eWMD$_{test}$ & eWMD$_{train}$  & NLL$_{test}$\\
    \midrule
    MLE & 1.015 \tiny{$\sigma$=0.023} & 0.947 \tiny{$\sigma$=0.019} & 2.365 \\
    SeqGAN & 2.900 \tiny{$\sigma$=0.025} & 3.118 \tiny{$\sigma$=0.018} & 3.122 \\
    RankGAN & 4.451 \tiny{$\sigma$=0.083} & 4.829 \tiny{$\sigma$=0.021} & 3.083\\
    MaliGAN & 4.891 \tiny{$\sigma$=0.061} & 4.962 \tiny{$\sigma$=0.020} & 3.240 \\
    LeakGAN & 1.803 \tiny{$\sigma$=0.027} & 1.767 \tiny{$\sigma$=0.023} & 2.327\\
    \cmidrule{1-4}
    CoT-basic& \textbf{0.766 \tiny{$\sigma$=0.031}} & \textbf{0.886 \tiny{$\sigma$=0.019}} & 2.247\\
    CoT-strong & 0.923 \tiny{$\sigma$=0.018} & 0.941 \tiny{$\sigma$=0.016} & \textbf{2.144}\\
    \bottomrule
   \end{tabular}
    \end{sc}
    \end{small}
    \end{center} 
\end{table} 
The results are shown in Table~\ref{tb:real_quality} and Table~\ref{tb:real_diversity}. 
In terms of generative quality, CoT-basic achieves state-of-the-art performance over all the baselines with the same architecture-level capacity, especially the long-term robustness at n-gram level. CoT-strong using a conservative generation strategy, \emph{i.e.} setting the inverse temperature parameter $\alpha$ higher than 1, as in \citep{guo2017long} achieves the best performance over all compared models. In terms of generative diversity, the results show that our model achieves the state-of-the-art performance on all metrics including NLL$_{test}$, which is the optimization target of MLE.

\noindent \textbf{Implementation Details of eWMD}
To calculate eWMD, we adopted a multi-layer convolutional neural network as the feature extractor. We calculate the gradient \emph{w.r.t.} the one-hot representation $O_s$ of the sequence $s$ for gradient penalty. The training loss of the Wasserstein critic $f_\omega$ can be formulated as
\begin{align*}
    L_c(\omega, \lambda) =& \mathop{\mathbb{E}}_{s \sim G_\theta} \left[f_\omega(O_s)\right] - \mathop{\mathbb{E}}_{s \sim p_{\text{data}}} \left[f_\omega(O_s)\right]\\&+ \lambda \max(0, \Vert \nabla f_\omega (\hat{O})\Vert_2 - 1)^2,
\end{align*}
where 
$$\hat{O} = (1 - \mu)O_{s_p}  + \mu O_{s_q}$$
$$\mu \sim \textbf{Uniform}(0,1)$$
$$s_q \sim G_\theta$$
$$s_p \sim p_{\text{data}}.$$
We use Adam \citep{kingma2014adam} as the optimizer, with hyper-parameter settings of $\alpha=1e-4$, $\beta_1=0.5$, $\beta_2=0.9$. For each evaluated generator, we train the critic $f_\omega$ for 100,000 iterations, and calculate eWMD($p_{\text{data}}, G_\theta$) as
$$\mathop{\mathbb{E}}_{s \sim p_{\text{data}}} \left[f_\omega(O_s)\right] - \mathop{\mathbb{E}}_{s \sim G_\theta} \left[f_\omega(O_s)\right].$$
The network architecture for $f_\omega$ is shown in Table~\ref{tb:cnn_archi}.

\begin{table}[h]
  \caption{Detailed implementation of eWMD network architecture.}\label{tb:cnn_archi}
  \begin{center}
  \begin{small}
  \begin{tabular}{c}
    \toprule
    Word Embedding Layer, hidden dim $=128$\\
    \midrule
    Conv1d, window size$=2$, strides$=1$, channels $=64$\\
    \midrule
    Leaky ReLU Nonlinearity ($\alpha=0.2$)\\
    \midrule
    Conv1d, window size$=3$, strides$=2$, channels $=64$\\
    \midrule
    Leaky ReLU Nonlinearity ($\alpha=0.2$)\\
    \midrule
    Conv1d, window size$=3$, strides$=2$, channels $=128$\\
    \midrule
    Leaky ReLU Nonlinearity ($\alpha=0.2$)\\
    \midrule
    Conv1d, window size$=4$, strides$=2$, channels $=128$\\
    \midrule
    Leaky ReLU Nonlinearity ($\alpha=0.2$)\\
    \midrule
    Flatten\\
    \midrule
    Fully Connected, output dimension $=512$\\
    \midrule
    Leaky ReLU Nonlinearity ($\alpha=0.2$)\\
    \midrule
    Fully Connected, output dimension $=1$\\
    \bottomrule
   \end{tabular}
    \end{small}
    \end{center} 
\end{table}

\section{Future Work \& Conclusion}
We proposed Cooperative Training, a novel algorithm for training generative models of discrete data. 
CoT achieves independent success without the necessity of pre-training via maximum likelihood estimation or involving REINFORCE.
In our experiments, CoT achieves superior performance on sample quality, diversity, as well as training stability.

As for future work, one direction is to explore whether there is better way to factorize the dropped term of Eq.~(\ref{eq:simplified_g_grad}) into some low-variance term plus another high-variance residual term. This would further improve the performance of models trained via CoT.
Another interesting direction is to investigate whether there are feasible factorization solutions for the optimization of other distances/divergences, such as Wasserstein Distance, total variance and other task-specific measurements.

\section{Acknowledgement}
The corresponding authors Sidi Lu and Weinan Zhang thank the support of National Natural Science Foundation of China (61702327, 61772333, 61632017), Shanghai Sailing Program (17YF1428200).

\bibliography{cot}

\begin{thebibliography}{19}
\providecommand{\natexlab}[1]{#1}
\providecommand{\url}[1]{\texttt{#1}}
\expandafter\ifx\csname urlstyle\endcsname\relax
  \providecommand{\doi}[1]{doi: #1}\else
  \providecommand{\doi}{doi: \begingroup \urlstyle{rm}\Url}\fi

\bibitem[Arjovsky \& Bottou(2017)Arjovsky and Bottou]{arjovsky2017towards}
Arjovsky, M. and Bottou, L.
\newblock Towards principled methods for training generative adversarial
  networks.
\newblock \emph{arXiv preprint arXiv:1701.04862}, 2017.

\bibitem[Arjovsky et~al.(2017)Arjovsky, Chintala, and
  Bottou]{arjovsky2017wasserstein}
Arjovsky, M., Chintala, S., and Bottou, L.
\newblock Wasserstein gan.
\newblock \emph{arXiv:1701.07875}, 2017.

\bibitem[Bahdanau et~al.(2014)Bahdanau, Cho, and Bengio]{bahdanau2014neural}
Bahdanau, D., Cho, K., and Bengio, Y.
\newblock Neural machine translation by jointly learning to align and
  translate.
\newblock \emph{arXiv:1409.0473}, 2014.

\bibitem[Bengio et~al.(2015)Bengio, Vinyals, Jaitly, and
  Shazeer]{bengio2015scheduled}
Bengio, S., Vinyals, O., Jaitly, N., and Shazeer, N.
\newblock Scheduled sampling for sequence prediction with recurrent neural
  networks.
\newblock In \emph{NIPS}, pp.\  1171--1179, 2015.

\bibitem[Che et~al.(2017)Che, Li, Zhang, Hjelm, Li, Song, and
  Bengio]{che2017maximum}
Che, T., Li, Y., Zhang, R., Hjelm, R.~D., Li, W., Song, Y., and Bengio, Y.
\newblock Maximum-likelihood augmented discrete generative adversarial
  networks.
\newblock \emph{arXiv:1702.07983}, 2017.

\bibitem[Goodfellow et~al.(2014)Goodfellow, Pouget-Abadie, Mirza, Xu,
  Warde-Farley, Ozair, Courville, and Bengio]{goodfellow2014generative}
Goodfellow, I., Pouget-Abadie, J., Mirza, M., Xu, B., Warde-Farley, D., Ozair,
  S., Courville, A., and Bengio, Y.
\newblock Generative adversarial nets.
\newblock In \emph{NIPS}, pp.\  2672--2680, 2014.

\bibitem[Gulrajani et~al.(2017)Gulrajani, Ahmed, Arjovsky, Dumoulin, and
  Courville]{gulrajani2017improved}
Gulrajani, I., Ahmed, F., Arjovsky, M., Dumoulin, V., and Courville, A.~C.
\newblock Improved training of wasserstein gans.
\newblock In \emph{NIPS}, pp.\  5769--5779, 2017.

\bibitem[Guo et~al.(2017)Guo, Lu, Cai, Zhang, Yu, and Wang]{guo2017long}
Guo, J., Lu, S., Cai, H., Zhang, W., Yu, Y., and Wang, J.
\newblock Long text generation via adversarial training with leaked
  information.
\newblock \emph{arXiv:1709.08624}, 2017.

\bibitem[Kingma \& Ba(2014)Kingma and Ba]{kingma2014adam}
Kingma, D.~P. and Ba, J.
\newblock Adam: A method for stochastic optimization.
\newblock \emph{arXiv preprint arXiv:1412.6980}, 2014.

\bibitem[Kusner et~al.(2015)Kusner, Sun, Kolkin, and
  Weinberger]{kusner2015word}
Kusner, M., Sun, Y., Kolkin, N., and Weinberger, K.
\newblock From word embeddings to document distances.
\newblock In \emph{International Conference on Machine Learning}, pp.\
  957--966, 2015.

\bibitem[Lamb et~al.(2016)Lamb, GOYAL, Zhang, Zhang, Courville, and
  Bengio]{lamb2016professor}
Lamb, A.~M., GOYAL, A. G. A.~P., Zhang, Y., Zhang, S., Courville, A.~C., and
  Bengio, Y.
\newblock Professor forcing: A new algorithm for training recurrent networks.
\newblock In \emph{NIPS}, pp.\  4601--4609, 2016.

\bibitem[Lin et~al.(2017)Lin, Li, He, Zhang, and Sun]{lin2017adversarial}
Lin, K., Li, D., He, X., Zhang, Z., and Sun, M.-T.
\newblock Adversarial ranking for language generation.
\newblock In \emph{NIPS}, pp.\  3155--3165, 2017.

\bibitem[Lu et~al.(2018)Lu, Zhu, Zhang, Wang, and Yu]{lu2018neural}
Lu, S., Zhu, Y., Zhang, W., Wang, J., and Yu, Y.
\newblock Neural text generation: Past, present and beyond.
\newblock \emph{arXiv preprint arXiv:1803.07133}, 2018.

\bibitem[Radford et~al.(2015)Radford, Metz, and
  Chintala]{radford2015unsupervised}
Radford, A., Metz, L., and Chintala, S.
\newblock Unsupervised representation learning with deep convolutional
  generative adversarial networks.
\newblock \emph{arXiv preprint arXiv:1511.06434}, 2015.

\bibitem[Ranzato et~al.(2015)Ranzato, Chopra, Auli, and
  Zaremba]{ranzato2015sequence}
Ranzato, M., Chopra, S., Auli, M., and Zaremba, W.
\newblock Sequence level training with recurrent neural networks.
\newblock \emph{arXiv preprint arXiv:1511.06732}, 2015.

\bibitem[Sutton(1984)]{sutton1984temporal}
Sutton, R.~S.
\newblock Temporal credit assignment in reinforcement learning.
\newblock 1984.

\bibitem[Ulyanov et~al.(2016)Ulyanov, Vedaldi, and
  Lempitsky]{ulyanov2016instance}
Ulyanov, D., Vedaldi, A., and Lempitsky, V.
\newblock Instance normalization: The missing ingredient for fast stylization.
\newblock \emph{arXiv preprint arXiv:1607.08022}, 2016.

\bibitem[Yu et~al.(2017)Yu, Zhang, Wang, and Yu]{yu2017seqgan}
Yu, L., Zhang, W., Wang, J., and Yu, Y.
\newblock Seqgan: Sequence generative adversarial nets with policy gradient.
\newblock In \emph{AAAI}, pp.\  2852--2858, 2017.

\bibitem[Zhu et~al.(2018)Zhu, Lu, Zheng, Guo, Zhang, Wang, and
  Yu]{zhu2018texygen}
Zhu, Y., Lu, S., Zheng, L., Guo, J., Zhang, W., Wang, J., and Yu, Y.
\newblock Texygen: A benchmarking platform for text generation models.
\newblock \emph{arXiv:1802.01886}, 2018.

\end{thebibliography}
\bibliographystyle{icml2019}



\section*{Supplementary material}

## A  Detailed Derivation of the Algorithm

$$(11) = \left( -\frac{1}{2} \mathop{\mathbb{E}}_{s_t \sim G_\theta} \left[ \log G_\theta(s_t) - \log M_\phi(s_t) \right] \right)$$

$$= \left( -\frac{1}{2} \mathop{\mathbb{E}}_{s_t \sim G_\theta} \left[ \frac{G_\theta(s_{t-1}) G_\theta(s_t|s_{t-1})}{G_\theta(s_t)} \left( \log G_\theta(s_t) - \log M_\phi(s_t) \right) \right] \right)$$

$$= \left( -\frac{1}{2} \mathop{\mathbb{E}}_{s_t \sim G_\theta} \left[ \frac{G_\theta(s_{t-1}) G_\theta(s_t|s_{t-1})}{G_\theta(s_t)} \left( \log G_\theta(s_t|s_{t-1}) G_\theta(s_{t-1}) - \log M_\phi(s_t|s_{t-1}) M_\phi(s_{t-1}) \right) \right] \right)$$

$$= -\frac{1}{2} \left( \sum_{s_t} G_\theta(s_{t-1}) G_\theta(s_t|s_{t-1}) \left( \log G_\theta(s_t|s_{t-1}) - \log M_\phi(s_t|s_{t-1}) \right) \right.$$
$$\left. + \sum_{s_t} G_\theta(s_{t-1}) G_\theta(s_t|s_{t-1}) \log \frac{G_\theta(s_{t-1})}{M_\phi(s_{t-1})} \right)$$

$$= -\frac{1}{2} \left( \sum_{s_t} G_\theta(s_{t-1}) G_\theta(s_t|s_{t-1}) \left( \log G_\theta(s_t|s_{t-1}) - \log M_\phi(s_t|s_{t-1}) \right) \right.$$
$$\left. + \sum_{s_{t-1}} \left( G_\theta(s_{t-1}) \log \frac{G_\theta(s_{t-1})}{M_\phi(s_{t-1})} \right) \sum_{s_t} G_\theta(s_t|s_{t-1}) \right) \quad \text{\small (here } s_{t-1} \text{ iterates over all prefixes of the sequences in } \{s_t\} \text{)}$$

$$= -\frac{1}{2} \left( \sum_{s_t} G_\theta(s_{t-1}) G_\theta(s_t|s_{t-1}) \left( \log G_\theta(s_t|s_{t-1}) - \log M_\phi(s_t|s_{t-1}) \right) + \sum_{s_{t-1}} G_\theta(s_{t-1}) \log \frac{G_\theta(s_{t-1})}{M_\phi(s_{t-1})} \right)$$

$$= -\frac{1}{2} \left( \sum_{s_t} G_\theta(s_{t-1}) G_\theta(s_t|s_{t-1}) \left( \log G_\theta(s_t|s_{t-1}) - \log M_\phi(s_t|s_{t-1}) \right) + \mathop{\mathbb{E}}_{s_{t-1} \sim G_\theta} \left[ \log \frac{G_\theta(s_{t-1})}{M_\phi(s_{t-1})} \right] \right)$$

$$= -\frac{1}{2} \left( \sum_{s_{t-1}} G_\theta(s_{t-1}) \sum_{s_t} G_\theta(s_t|s_{t-1}) \left( \log G_\theta(s_t|s_{t-1}) - \log M_\phi(s_t|s_{t-1}) \right) + \mathop{\mathbb{E}}_{s_{t-1} \sim G_\theta} \left[ \log \frac{G_\theta(s_{t-1})}{M_\phi(s_{t-1})} \right] \right)$$

$$= (12)$$

## B  Sample Comparison and Discussion

Table 1 shows samples from some of the most powerful baseline models and our model.

Observation of the model samples indicates that:

- CoT produces remarkably more diverse and meaningful samples when compared to Leak-GAN.
- The consistency of CoT is significantly improved when compared to MLE.

## C  Further Discussions about the Experiment Results

**The Optimal Balance for Cooperative Training**  We find that the same learning rate and iteration numbers for the generator and mediator seems to be the most competitive choice. As for the architecture choice, we find that the mediator needs to be slightly stronger than the generator. For the best result in the synthetic experiment, we adopt exactly the same generator as other compared models and a mediator whose hidden state size is twice larger (with 64 hidden units) than the generator.

Theoretically speaking, we can and we should sample more batches from $G_\theta$ and $P$ respectively for training the mediator in each iteration. However, if no regularizations are used when training the mediator, it can easily over-fit, leading the generator's quick convergence in terms of $KL(G_\theta \| P)$ or $\text{NLL}_{oracle}$, but divergence in terms of $JSD(G_\theta \| P)$. Empirically, this could be alleviated by applying dropout techniques (Srivastava et al., 2014) with 50% keeping ratio before the output layer of RNN. After applying dropout, the empirical results show good consistency with our theory that, more training batches for the mediator in each iteration is always helpful.

However, applying regularizations is not an ultimate solution and we look forward to further theoretical investigation on better solutions for this problem in the future.

Table 1: WMT News Samples from Different Models

| Sources | Example |
|---------|---------|
| LeakGAN | (1) It's a big advocate for therapy is a second thing to do, and I'm creating a relationship with a nation. |
| | (2) It's probably for a fantastic footage of the game, but in the United States is already time to be taken to live. |
| | (3) It's a sad House we have a way to get the right because we have to go to see that, " she said. |
| | (4) I'm not sure if I thank a little bit easier to get to my future commitment in work, " he said. |
| | (5) " I think it was alone because I can do that, when you're a lot of reasons, " he said. |
| | (6) It's the only thing we do, we spent 26 and $35(see how you do is we lose it," said both sides in the summer. |
| CoT | (1) We focus the plans to put aside either now, and which doesn't mean it is to earn the impact to the government rejected. |
| | (2) The argument would be very doing work on the 2014 campaign to pursue the firm and immigration officials, the new review that's taken up for parking. |
| | (3) This method is true to available we make up drink with that all they were willing to pay down smoking. |
| | (4) The number of people who are on the streaming boat would study if the children had a bottle - but meant to be much easier, having serious ties to the outside of the nation. |
| | (5) However, they have to wait to get the plant in federal fees and the housing market's most valuable in tourism. |
| MLE | (1) after the possible cost of military regulatory scientists, chancellor angela merkel's business share together a conflict of major operators and interest as they said it is unknown for those probably 100 percent as a missile for britain. |
| | (2) but which have yet to involve the right climb that took in melbourne somewhere else with the rams even a second running mate and kansas. |
| | (3) " la la la la 30 who appeared that themselves is in the room when they were shot her until the end " that jose mourinho could risen from the individual . |
| | (4) when aaron you has died, it is thought if you took your room at the prison fines of radical controls by everybody, if it's a digital plan at an future of the next time. |

**Possible Derivatives of CoT** The form of equation 11 can be modified to optimize other objectives. One example is the backward KLD (*a.k.a.* Reverse KLD) *i.e.* $KL(G\|P)$. In this case, the objective of the so-called "Mediator" and "Generator" thus becomes:

"Mediator", now it becomes a direct estimator $\hat{P}_\phi$ of the target distribution $P$:

$$J_{\hat{p}}(\phi) = \mathop{\mathbb{E}}_{s \sim P}[-\log(\hat{P}_\phi(s))]. \tag{1}$$

Generator:

$$J_g(\theta) = \sum_{t=0}^{n-1} \mathop{\mathbb{E}}_{s_t \sim G_\theta} \left[ \pi_g(s_t)^\top (\log \pi_{\hat{p}}(s_t) - \log \pi_g(s_t)) \right]. \tag{2}$$

Such a model suffers from so-called mode-collapse problem, as is analyzed in Ian's GAN Tutorial (Goodfellow, 2016). Besides, as the distribution estimator $\hat{P}\phi$ inevitably introduces unpredictable behaviors when given unseen samples *i.e.* samples from the generator, the algorithm sometimes fails (numerical error) or diverges.

In our successful attempts, the algorithm produces similar (not significantly better than) results as CoT. The quantitive results are shown as follows:

Table 2: N-gram-level quality benchmark: BLEU on test data of EMNLP2017 WMT News (New Split)

| Model/Algorithm | BLEU-2 | BLEU-3 | BLEU-4 | BLEU-5 | eWMD |
|-----------------|--------|--------|--------|--------|------|
| CoT-basic (ours) | 0.850 | 0.571 | 0.316 | 0.169 | **1.001** ($\sigma = 0.020$) |
| Reverse KL (ours) | **0.860** | **0.590** | **0.335** | **0.181** | 1.086 ($\sigma = 0.014$) |

Although under evaluation of weak metrics like BLEU, if successfully trained, the model trained via Reverse KL seems to be better than that trained via CoT, the disadvantage of Reverse KL under evaluation of more strict metric like eWMD indicates that Reverse KL does fail in learning some aspects of the data patterns *e.g.* completely covering the data mode.

## REFERENCES

Ian Goodfellow. Nips 2016 tutorial: Generative adversarial networks. *arXiv preprint arXiv:1701.00160*, 2016.

Nitish Srivastava, Geoffrey Hinton, Alex Krizhevsky, Ilya Sutskever, and Ruslan Salakhutdinov. Dropout: A simple way to prevent neural networks from overfitting. *The Journal of Machine Learning Research*, 15(1):1929–1958, 2014.



\end{document}